\documentclass[a4paper,11pt]{article}

\usepackage{graphicx}  
\usepackage{url}       
\usepackage{times}
\usepackage{natbib}
\usepackage[margin=25mm]{geometry}
\usepackage{booktabs}
\usepackage{microtype}

\title{Role Semantics for Better Models of Implicit Discourse Relations}

\date{}

\author{Michael Roth\\
       Department of Language Science and Technology, Saarland University\\
       \texttt{mroth@coli.uni-sb.de}
}

\begin{document}
\maketitle
\thispagestyle{empty}
\pagestyle{empty}

\newcommand\BibTeX{B{\sc ib}\TeX}

\begin{abstract}
\noindent
Predicting the structure of a discourse is challenging because relations between discourse segments are often implicit and thus hard to distinguish computationally. I extend previous work to classify implicit discourse relations by introducing a novel set of features on the level of semantic roles.
My results demonstrate that such features are helpful, yielding results competitive with other feature-rich approaches on the PDTB. My main contribution is an analysis of improvements that can be traced back to role-based features, providing insights into why and when role semantics is helpful.

\end{abstract}

\section{Introduction}

Understanding natural language texts involves, inter alia, correctly identifying coherent segments and the relations that hold between them. Recognizing discourse relations is an important part of this process because such relations not only conceptualize which parts of a text belong together but also \textit{how} they are related. Apart from direct applications in text analysis (e.g., discourse parsing), recognizing discourse relations has further proven a useful preprocessing step for a range of downstream tasks \cite[inter alia]{louis10a,guzman14,narasimhan15,chandrasekaran17}. 

From a computational perspective, it has been shown that recognizing discourse relations can be performed with high accuracy when explicit discourse markers are available \cite[]{pitler08b}. However, classifying relations without explicit markers, so-called \textit{implicit} discourse relations, has persisted as a difficult task to date (cf.\ Xue et al., 2016\nocite{xue16}). One of the main challenges, as identified in \cite{lin09}, is the need to perform inference over two discourse segments. In this paper, I propose a new set of features based on semantic roles to address this challenge.
These features are meant to provide a shallow form of semantic representation, which might help a classifier to make better informed classification decisions. I argue that role semantic representations are particularly well-suited for this task because different types of discourse relations are defined over the propositions that they connect. For example, definitions in the Penn Discourse TreeBank 2.0 annotation manual \cite[]{prasad07} explicitly refer to role-level concepts such as events, situations and involved participants. In Rhetorical Structure Theory \cite[]{mann88}, some definitions contain references to concepts akin to \textit{proto-roles} (e.g. ``someone's deliberate action''). 
To illustrate the usefulness of role semantics for the classification of implicit discourse relations, consider the two sentenes shown in Example~(1):

\begin{itemize}
\item[(1) a.] ``Mr.\ Brady phoned Mr.\ Greenspan, \ldots'' \\[-2em]
\item[b.] ``He continued to work the phones through the weekend.''\\[-2em]
\item[] Relation: then, \textit{Temporal.Asynchronous.Precedence} \hfill (source: \texttt{wsj\_2413.pdtb})
\end{itemize}

In terms of frame-semantic representation \cite[]{fillmore76}, the roles involved in the second sentence can be identified as an \texttt{Ongoing\_activity} (the argument of ``continue''), a definite \texttt{Duration} and a pronominal \texttt{Agent}.\footnote{Roles based on FrameNet, see \url{http://framenet.icsi.berkeley.edu/}.} These cues indicate a sequence of situations with the same actor, making it likely that a \textit{Temporal} relation holds to the previous sentence.


\section{Discourse Relation Classification with Feature-rich Models}
\label{sec:model}

The task addressed in this paper is to determine the discourse relations that hold between two implicitly related discourse segments. In this section, I introduce a combined model for this task that aggregates outputs from multiple simpler models (\ref{subsec:model}), each of which uses only one type of feature. I then introduce new feature sets based on semantic roles (\ref{subsec:rolefeats}).

\subsection{Model and Previous Features}
\label{subsec:model}

My motivation for a model combination derives from the observation that
different types of features 
from the literature greatly vary with respect to the associated number of
feature instances and how well they generalize.
Consequently, there is no unique set of hyperparameters (e.g.\ level of
regularization, thresholding) that works best for all feature types. The proposed combined model consists of two steps to make use of information from inherently
diverse feature types. First, I train simple discourse relation classifiers
that only use one feature type each. Outputs from multiple classifiers are then
combined using averaging as a simple but effective form of model
combination.\footnote{Sum/averaging is used here because of its simplicity and robustness \cite[]{kittler98}. Due to the small development set size, methods with additional parameters may tend to overfit.}

I formalize the classification
of an instance~$i$ with respect to a discourse relation $r$ as follows. Given a
set of $n$ feature types, feature values are extracted and a set of
simple classifiers $c_{r,1}\ldots c_{r,n}$ are trained. At test time, each classifier outputs an individual
score $\mathit{score}_{c_{r,j}}(i)\in [0,1]$.
Decisions of multiple classifiers are then aggregated
by computing the arithmetic mean of the individual scores. 
As single classifiers, I use logistic regression models with L2 loss, as  implemented in the LIBLINEAR toolkit \cite[]{lee15}. Accordingly, the aggregated model predicts a relation $r$ for instance $i$ iff $\frac{1}{n} \Sigma_{j=1...n} \mathit{score}_{c_{r,j}}(i)>0.5$.

The following list provides an overview of all feature sets from the literature that I reimplemented for the described approach, and gives the total number of features for each type.

\paragraph{First/Last.} Set of indicators for the first and last words in each discourse segment. In case of Example~(1), instances of this feature set include \texttt{1:FIRST:Mr.}, \texttt{2:FIRST:He}, etc. (for details, see Pitler et al., 2009\nocite{pitler09}). \hfill ca.~$74\,000$~features

\paragraph{Dates and number.} Indicator features for the number of date and number expressions in each discourse segment (e.g. \texttt{1:DATE:0}; see Pitler et al., 2009\nocite{pitler09}). \hfill ca.~$10\,000$

\paragraph{Production rules.} Features on production rules used to construct each discourse segment's constituency tree (e.g. \texttt{1:S\_NP\_VP}; see Lin et al., 2009\nocite{lin09}). \hfill ca.~$78\,000$

\paragraph{Verb features.} Indicators for the main verb, its tense/modality and average verb phrase length~(e.g. \texttt{1:VERB:phone}, \texttt{2:TENSE:past}; see Park and Cardie, 2012\nocite{park12}).\hfill ca.~$20\,000$

\paragraph{Coreference.} Set of features that indicate coreferring mentions, as predicted by Stanford CoreNLP \cite[]{lee13}, across two related discourse segments (see Rutherford and Xue, 2014\nocite{rutherford14}).\hfill ca.~$10\,000$

\paragraph{Brown clusters.} Feature sets indicating precomputed Brown cluster IDs \cite[]{turian10} of words occurring in each discourse segment (e.g. \texttt{2:11100110}; see Braud and Denis, 2015\nocite{braud15}). \mbox{\quad} \hfill $200$--$6\,400$ 

\paragraph{Pairwise Brown clusters.} Pairwise Brown cluster IDs indicating word pairs across two related discourse segments (e.g. \texttt{11110110x11000100}; see Braud and Denis, 2015\nocite{braud15}). \mbox{\quad} \hfill 
up~to~$10$~million

\subsection{Features based on Semantic Roles}
\label{subsec:rolefeats}

As new features, I propose to utilize the semantic roles identified in a pair
of discourse segments. I define two variants of this feature type: one based on
FrameNet \cite[]{ruppenhofer10} and one based on PropBank \cite[]{palmer05}. All
features are computed automatically using a state-of-the-art semantic role
labeler \cite[]{roth16,rothlapata16}. Each variant includes both raw labels as
well as a combination of the label and the filler word to which the label is
assigned. To reduce sparsity, filler words are always represented by pre-computed Brown cluster IDs \cite[]{turian10}. The list below provides additional details as well as example instances based on the sentences shown in Example~(1).

\paragraph{FrameNet roles.} This feature set indicates all frame elements that are identified in a pair of related discourse segments. For instance, two frame element fillers are identified in the phrase \textit{he continued to work}: \textit{he} is the \texttt{Agent} of the frame evoked by the verb \textit{work}, and \textit{work} itself fills the \texttt{Ongoing\_activity} element of the frame evoked by \textit{continue}. To compute features for \textit{he}, the Brown cluster ID of the word is looked up (\texttt{11100110}) and it is determined that the word occurs in the \texttt{2}nd discourse segment in Example~(1). Accordingly, the indicator features that represent \textit{he} and its semantic role in this case are \texttt{2:Agent} and \texttt{2:Agent:11100110}.\footnote{I also experimented with feature conjunctions in order to explicitly model semantic interactions between two discourse segments. However, such conjuctions consistently reduced development performance, probably due to sparsity.} \hfill ca.~$37\,000$

\vspace{-1em}
\paragraph{PropBank roles.} Analogous to the FrameNet features, this feature set consists of indicators for PropBank labels. Because argument labels in PropBank (A0\ldots A5) are only meaningful with respect to a given predicate, I define two conjoined versions of this feature type: one takes into account the predicate's class in VerbNet \cite[]{kipper08} and one the predicate lemma itself (e.g., \texttt{2:work-73.2\_A0} and \texttt{2:work\_A0:11100110}, resp.). In each variant, predicate-independent labels (modifiers such as time and location) are optionally considered in the same representation format. \hfill ca.~$560\,000$

\section{Experiments}
\label{sec:exps}

I evaluate the proposed model on version 2.0 of the Penn Discourse Treebank
\cite[PDTB,][]{prasad08}. To ensure a fair comparison, I use the same preprocessing and
weighting techniques as well as the same data instances as previous work
\cite[]{rutherford14,braud15}. That is, each instance is a pair of implicitly related discourse segments as annotated in the PDTB corpus. Sections 2--20 of the corpus are used for training, 21--22 for testing, and all other sections for development. 

\paragraph{Baseline and comparison models.} I use three variants of the proposed model to directly examine the utility of semantic roles and combining classifiers. The first two models are instances of the feature-rich model described in Section~\ref{sec:model}, with hyperparameter tuning and feature selection done on the training and development sets: \textit{AverageFeats} uses a combination of feature sets described in subsection~\ref{subsec:model}, whereas \textit{AverageFeats+SRL} also uses the role-level features from subsection~\ref{subsec:rolefeats}. Note that for each type of role set at most one feature representation is chosen. All feature sets are selected based on the best performance on the development set. The third model, \textit{AllFeats}, is a baseline logistic regression classifier that uses all best development feature sets at the same time.

For comparison, I consider a range of current state-of-the-art models. The best feature-rich models \cite[]{rutherford14,braud15} use a range of binary indicator features largely
identical to the features described in Section~\ref{subsec:model}. The
most notable difference to this work is that Rutherford and Xue use a small list of coreference patterns in addition to features that simply indicate coreferring mention counts. Neural-network models \cite[]{zhang15,liu16b,qin16} use attention or convolution mechanisms to identify important words and word spans in each discourse segment. They then predict the discourse relation based on a composition function applied over representations of important words. All of the comparison models use the same training and test instances as this work and are directly comparable.


\begin{table}[!htb]
    \begin{minipage}{.48\linewidth}
\begin{tabular}{@{}l@{ ~}c@{ ~}c@{ ~}c@{ ~}c@{}}
\toprule
\multicolumn{2}{r}{comp} & cont & exp & temp \\
\midrule
\multicolumn{5}{l}{\textbf{Neural network models}}\\
\cite{zhang15}                                       & 33.2 & 52.0 & 69.6 & 30.5 \\
\cite{liu16b}										& 36.7 & 54.5 & 70.4 & \textbf{38.8} \\
\cite{qin16}											& \textbf{41.6} & \textbf{57.3} & \textbf{71.5} & 35.4 \\
\midrule
\multicolumn{5}{l}{\textbf{Recent feature-rich models}}\\
\cite{rutherford14}                          & \underline{39.7} & 54.4 & \underline{70.2} & 28.7 \\
\cite{braud15}                                       & 36.4 & 55.8 & 67.4 & 29.3 \\
\midrule
\multicolumn{5}{l}{\textbf{This work's models}}\\
\textit{AverageFeats}      & 36.3 & 55.9 & 69.4 & 30.5 \\
\textit{AverageFeats+SRL}  & 37.0 & \underline{56.3} & 69.4 & \underline{32.1} \\
\textit{AllFeats}          & 34.5 & 51.3 & 60.4 & 26.8     \\ 
\bottomrule
\end{tabular}
\caption{One-vs-all results in F$_1$-score on the four PDTB top-level relations (\textit{comp}arison, \textit{cont}ingency, \textit{exp}ansion and \textit{temp}oral). Best overall results are marked in bold, best results by feature-rich models are underlined.}
\label{tbl:results}
    \end{minipage}%
    \begin{minipage}{.04\linewidth}
	\quad
    \end{minipage}%
    \begin{minipage}{.47\linewidth}
\begin{tabular}{lc@{ ~ ~ ~ ~}r}
\\[0.25em]
\toprule
Role name & Position & Weight \\
\midrule
Request & segment 1 & $+1.13061$ \\
Addressee & segment 1 & $+0.90852$ \\
Relative\_time & segment 1 & $+0.85555$ \\
\midrule
Stuff & segment 2 & $+0.79267$ \\
Success\_or\_failure & segment 2 & $+0.69008$ \\
Unattr\_information & segment 2 & $+0.66578$ \\
\midrule
Agent & segment 1 & $+0.39992$ \\
Agent & segment 2 & $-0.68378$ \\
\bottomrule
\end{tabular}
\caption{List of indicator features on FrameNet frame elements that received a high weight for recognizing the discourse relation Contingency.}
\label{tbl:weights}
    \end{minipage} 
\end{table}

\paragraph{Results.} Table~\ref{tbl:results} lists F$_1$-scores for each of the
top-level relations in the PDTB test set. Note that multiple relation
types can apply to one relation instance. Hence, instead of one 4-way classification,
this task is traditionally separated into four binary tasks. The results show
that \textit{AverageFeats} performs competitively with other feature-rich models
for discourse relation classification. Additional features on semantic roles
improve performance for all but one relation. In the cases in which semantic roles are helpful, both FrameNet-based and PropBank-based feature sets are selected. Two of the four scores by
\textit{AverageFeats+SRL} represent the best reported results with a
feature-rich model. The performance of \textit{AllFeats} is consistently worse
than those of other recent models.
This complies with my hypothesis that hyperparameters tuned for one single
model do not generalize well across different feature types.

\paragraph{Discussion.} One advantage of simple classification models based on binary features is that predictions based on learned feature weights can easily be interpreted. In the following, I take a closer look at classification instances that the model \textit{AverageFeats+SRL} got correct but that were misclassified by the other models. The weights of the features that apply in these examples provide insights as to how and when semantic roles are beneficial. For simplicity, I focus the discussion on FrameNet roles (i.e.~frame element types). 

For the implicit relation \textit{Contingency}, the learned feature weights indicate that its prediction becomes more likely when an \texttt{Agent} is identified in the first discourse segment (high positive feature weight) but not in the second segment (negative feature weight). This seems to reflect the fact that most of these relations connect a cause and a result, as shown for instance in Example (2).

\begin{enumerate}
\item[(2)] ``\ldots traders can buy or sell even when they don't have a customer order \ldots [\textit{as a result}] liquidity becomes a severe problem for thinly traded contracts \ldots'' \hfill (\texttt{wsj\_2110.pdtb})
\end{enumerate}

Semantic roles are helpful in such cases because they provide a means to distinguish events initiated by someone (the cause) from simple states (the result). A list of features that seem to contribute to this distinction, as identified by their associated feature weights, are given in Table~\ref{tbl:weights}.

The feature weights assigned in role-based classifiers for other discourse
relations are overall smaller and thus harder to interpret. Still, certain
trends can be observed. For example, I find that co-occurrences of specific
roles in both connected discourse segments may indicate a
\textit{Comparison}. Example (3) shows one such instance, in which the role
\texttt{Purpose} has been identified in both segments (assigned feature weight:
$+0.117$). Other roles, for which the same pattern of weights are observed
include, among others, \texttt{Theme} ($+0.435$) and \texttt{Businesses}
($+0.254$).

\begin{enumerate}
\item[(3)] ``Her goal: to top 300 ad pages ... [\textit{but}] whether she can meet that ambitious goal is still far from certain.'' \hfill (\texttt{wsj\_2109.pdtb})
\end{enumerate}

Concerning the \textit{Temporal} relation, high feature weights are learned for specific FrameNet roles, such as \texttt{Activity\_start} in the first discourse segment ($+1.654$) and \texttt{Process\_end} in the second segment ($+1.116$). Even though these feature weights seem to be intuitive, they only lead to marginal improvements to the absolute classification performance, presumably because textual order in discourse not necessarily represents linear temporal order (``before'' vs.\ ``after''). Higher gains could be achieved if training and evaluation was performed on more specific relation annotations but such instances are too rare in practice for the feature-rich classifiers to learn robust generalizations: For example, the current version of the Penn Discourse Treebank contains a total of only 151 implicit relation instances of the discourse relation \textit{Temporal.Asynchronous.Succession}.

\section{Related Work}
\label{sec:relwork}

The task of predicting implicit discourse relations was first introduced in the context of implicit and explicit relation classification \cite[]{marcu02}. \cite{pitler09} were the first to address implicit relations specifically. They applied a Naive Bayes model with a range of binary features. Follow-up work examined different methods for feature selection \cite[]{lin09,park12} as well as novel feature types based on pairs of word classes/clusters, entity mentions, and word embeddings \cite[]{biran13,louis10b,braud15}. Further improvements were made via multi-task learning \cite[]{lan13} and training data expansion \cite[]{rutherford15}.

In recent years, a myriad of neural-network based models have been proposed for the task of recognizing implicit discourse relations \cite[inter alia]{ji14,zhang15,liu16b,qin17}. Models of this kind have a high expressive power and generally outperform methods that rely on manual feature engineering. However, being able to trace back improvements to individual features was key to my discussion in Section~\ref{sec:exps}. Recent results in downstream NLP tasks indicate that neural network models can perform better when incorporating binary features \cite[inter alia]{cheng16,sennrich16}. 

\section{Conclusions}
\label{sec:conclusions}

I proposed a simple model combination for discourse relation classification that aggregates outputs from multiple classifiers.
Several classifiers use novel features based on automatic semantic role labeling. I have shown that such features improve classification performance and provide shallow insights into relationships between role semantics and discourse semantics. 

In the future, I plan to apply more sophisticated methods of model ensembling. I would like to investigate whether neural network approaches to discourse relation classification can also benefit from  structural information in the form of semantic roles. I believe this to be a promising research direction especially because of the small size of available training data, which presumably makes it difficult for a neural network to learn any higher level structures by itself.

\section*{Acknowledgements}

This research was supported in part by the Cluster  of  Excellence  ``Multimodal  Computing  and Interaction''  of  the  German  Excellence  Initiative, and by a DFG Research Fellowship (RO 4848/1-1).

\bibliography{new}
\bibliographystyle{chicago}

\end{document}